\documentclass[lettersize,journal]{IEEEtran}
\usepackage{amsmath,amsfonts}
\usepackage{algorithmic}
\usepackage{algorithm}
\usepackage{array}
\usepackage[caption=false,font=normalsize,labelfont=sf,textfont=sf]{subfig}
\usepackage{textcomp}
\usepackage{stfloats}
\usepackage{url}
\usepackage{verbatim}
\usepackage{graphicx}
\usepackage{cite}

\usepackage{hhline}
\usepackage{multirow,amssymb}
\usepackage{hyperref}
\usepackage[table,xcdraw]{xcolor}

\usepackage{makecell}

\usepackage{cellspace}
\begin{document}

\title{RoME: Role-aware Mixture-of-Expert Transformer for Text-to-Video Retrieval}

\author{Burak Satar,~\IEEEmembership{Member,~IEEE,} Hongyuan Zhu, Hanwang Zhang, and Joo Hwee Lim,~\IEEEmembership{Senior Member,~IEEE,}
\thanks{Manuscript submitted June 26, 2022}}



\maketitle

\begin{abstract}
Seas of videos are uploaded daily with the popularity of social channels; thus, retrieving the most related video contents with user textual queries plays a more crucial role. Most methods consider only one joint embedding space between global visual and textual features without considering the local structures of each modality. Some other approaches consider multiple embedding spaces consisting of global and local features separately, ignoring rich inter-modality correlations.

We propose a novel mixture-of-expert transformer RoME that disentangles the text and the video into three levels; the roles of spatial contexts, temporal contexts, and object contexts.  
We utilize a transformer-based attention mechanism to fully exploit visual and text embeddings at both global and local levels with mixture-of-experts for considering inter-modalities and structures' correlations. The results indicate that our method outperforms the state-of-the-art methods on the YouCook2 and MSR-VTT datasets, given the same visual backbone without pre-training. Finally, we conducted extensive ablation studies to elucidate our design choices.

\end{abstract}


\section{Introduction}

Video content has become significantly abundant with the usage of mobile phones and social media. The rapid growth and the complex nature of videos have motivated various research such as video retrieval, visual question answering, video captioning, etc. Specifically, text-to-video retrieval aims to retrieve relevant video clips, given that a user text query becomes even more compelling since it is used daily by billions of users over various video-based applications. In this work, we focus only on text-to-video retrieval as the text allows semantic queries, which are commonly used to search for relevant videos.

Semantic mapping and embedding are necessary to align textual and video features to achieve reasonable text-to-video retrieval performance. The traditional models \cite{Chang2015SemanticCD, Habibian2014CompositeCD} are inadequate to retrieve compositional video clips since they make use of the textual queries only in the form of keywords. Thus, many recent works utilize fine-grained textual descriptions \cite{Chen_2020_CVPR, mithun2020, miech18learning,miech19howto100m, miech20endtoend}. However, these works apply only one joint embedding space for text and video matching, which causes the loss of fine-grained details \cite{Liu2019a, mithun2020, dong_cvpr19}. Although there are some current methods with multiple embedding spaces, they treat each space individually without considering their correlations. For instance, \cite{Chen_2020_CVPR} uses only global features and soft attention at every level, lacking interaction between visual features. \cite{gabeur2020mmt} uses only self-attention to model correlations between visual features, which would make the model computationally expensive.

We propose a novel Role-aware Mixture-of-Expert transformer model (RoME) for the text-to-video retrieval task by exploiting different experts from textual and visual representations. This paper is an extension of our ICIP paper \cite{satar_2021} with the following summary of the innovations: 

1) Firstly, as in our conference version \cite{satar_2021}, we also disentangle the video and textual features into hierarchical structures with the motivation from various cognitive science studies \cite{Christopher, Zacks2001PerceivingRA} showing that people perceive events hierarchically from the object, scene, and event. Then, we calculate the similarities between the corresponding embedding levels in the text and video.

2) Secondly, different from our conference version, which applies the same attention scheme at both global and local level features, we only use inter-modality attention for the global level features which depict scenes and events, while we use intra- and inter-modality attention for the local level features, which contains fine-grained action and object. The intuition behind this is that global features are complementary to local representations through contexts; however, the local features are usually noise which can influence the global representations if applied symmetrically. 

3) Thirdly, we strengthen our model by applying the weighted sum to the video encodings, which is better than the common approach of simply averaging the embeddings. While the previous approach assumes that encoding level is equally important, our results show that contribution of each level is different and could change based on the dataset.  

4) Lastly, we conduct a more comprehensive experiment and outperform all the SOTA methods at YouCook2 and MSR-VTT datasets when using the same visual features to demonstrate our method's strong performance. 

\begin{figure*}[htb]
\centering
\centerline{\includegraphics[width=\textwidth]{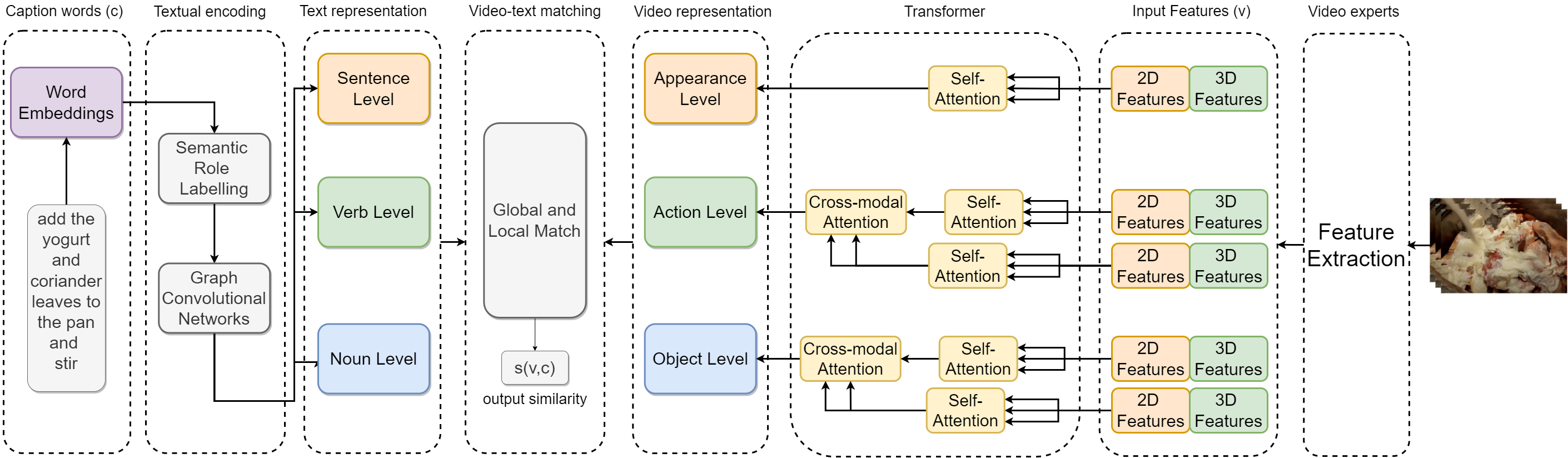}}
\caption{ Illustration of our model for video retrieval task. The input contains the extracted features from the video clips and their corresponding textual captions in training. The caption is disentangled into three levels in the textual encoding part through semantic role labeling and GCNs. In the video encoding part, the video expert features, namely 2D and 3D, are exploited by various transformers. While we use only self-attention at the global level, we utilize both self-attention and cross-modal attention at the local levels. We apply text-video representation matching as the last step. The weighted sum of visual experts is used to calculate similarities for the three-level embeddings. }\vspace{-10pt}
\label{fig:model}
\end{figure*}

\section{Related Work}
For images, the technique by Lee \textit{et al.} \cite{lee2018stacked} retrieves image-text by embedding visual regions and words in a shared space. Images with corresponding words and image regions show a high degree of resemblance. Karpathy \textit{et al.} \cite{10.1109/TPAMI.2016.2598339} deconstruct images and sentences into several areas and words in their proposal to use maximum alignment to calculate global matching similarity. And image descriptions are broken down into various roles by Wu \textit{et al.} \cite{8954449}. For videos, before the deep learning era, the researchers \cite{Chang2015SemanticCD, Habibian2014CompositeCD} used only textual keywords to retrieve short video clips with simple actions. However, it is insufficient to address video clips with complex actions since many works use natural language as queries. 

These works \cite{miech20endtoend,Rouditchenko2021AVLnetLA} typically applies only one embedding space to measure the gap between text and video. Although some of them use multimodal features from the video, having joint embedding space causes the loss of the fine-grained. For instance, Mithun \textit{et al.} \cite{mithun2020} and Liu \textit{et al.} \cite{Liu2019a}  utilize multimodal features, including image, motion, and audio. Dong \textit{et al.} \cite{dong_cvpr19} use mean pooling, biGRU, and CNN as three branches to encode consecutive videos and texts. In addition to this semantic gap, Luo \textit{et al.} \cite{Luo2020UniVL} use only self-attentions among all modalities, and their method is highly dependent on pre-training. Kim \textit{et al.} \cite{swamp} use only soft-attentions and ignore interaction between visual features. While Yang \textit{et al.} \cite{taco} utilize a fine-grained alignment with an extra loss and exploit special arrangement for hard negatives, they still use only self-attentions among all modalities by ignoring the interaction between visual features. Various other works focused on aligning local features to close the semantic gap \cite{song2019polysemous, Yu_2018_ECCV}, however, with partial progress to the problem. The sequential interaction of videos and texts is proposed to be combined by Yu \textit{et al.} \cite{Yu_2018_ECCV}.

Various methods \cite{videobert, Zhu_2020_Actbert, wang-etal-2020-vd} apply BERT models to the vision encoders for better video and text matching. For instance, Sun \textit{et al.} \cite{videobert} transform visual features into visual tokens and feed them into the network along with textual tokens. However, visual tokens may not handle the fine-grained details. To address this, Zhu \textit{et al.} \cite{Zhu_2020_Actbert} uses local visual features in a similar BERT-like network by ignoring global features. Rather than transformer-based approaches, some studies such as Tan \textit{et al.} \cite{tan2020wman} utilize LSTM and GNNs to exploit cross-modal features. Liu \textit{et al.} \cite{Liu2019a} utilizes only a gating mechanism to modulate embeddings as an alternative. However, it is challenging to capture high-level inter-modality information. Moreover, pre-training is used to bring a boost to the results. While a giant instructional dataset \cite{miech19howto100m} is common among the community, Alayrac \textit{et al.} \cite{alayrac2020selfsupervised} exploit another huge dataset on audio \cite{audioset}. Various studies focus on improving the quality of pre-training as well. While Gabeur \textit{et al.} \cite{mask_gabeur} mask their modalities respectively, Wang \textit{et al.} \cite{wang_object_pretraining} exploit object features for that aim. However, we are motivated on development without pre-training, although our work can benefit from pre-trained models with experimental results in Sec. IV.  

Some recent works have multiple embedding spaces and use both global/local features with various attention-based mechanisms. For example, Ging \textit{et al.} \cite{coot} exploit cross-attentions and self-attentions even though the interaction between visual features is limited. However, the action and object-related features are not considered. For fine-grained action retrieval, Wray separates action phrases into verbs and nouns, but this is difficult to apply to sentences with more intricate compositions. The hierarchical modeling of movies and paragraphs is proposed by Zhang \textit{et al.} \cite{zhang_hierarch}; however, it does not apply to the breakdown of single sentences. Gabeur \textit{et al.} \cite{gabeur2020mmt} use only self-attentions to learn the text-video correspondence even with many visual modalities. Also, having a large input size because of many visual expert features causes a high computation power due to the nature of transformers. Gabeur \textit{et al.} \cite{gabeur2020mmt} develop their model based on Miech \textit{et al.} \cite{miech18learning}. While the latter uses word2vec vectors and gated units, the former uses transformers in the textual and visual encoding part. Second, Gabeur \textit{et al.} exploit three more video experts than Miech \textit{et al.}. Chen \textit{et al.} \cite{Chen_2020_CVPR} uses only global 2D CNN features as a visual backbone and ignores interaction between visual feature levels. They suggest transforming textual and visual features into three semantic roles. Many works use various frozen video expert features as well. For instance, while \cite{Liu2019a, gabeur2020mmt} use seven video expert features, \cite{Gabeur2022Masking, miech18learning} use four of them. Our method could be named as a mixture of experts because we exploit the contextual interaction between the experts at every hierarchical level. These video experts could also be used in other vision-language-related tasks, such as video captioning \cite{MDVC_Iashin_2020} and VQA \cite{vlmo}.

Lastly, Our recent effort \cite{satar_2021} exploits fine-grained details in the global and local visual features, has multiple embedding spaces to close the semantic gap, and utilizes specially arranged transformers to make use of modality-specific and modality-complement features. However, our earlier work uses these transformers regardless of the hierarchical level. In other words, the same mechanisms are applied at every level. Moreover, the model simply averages the similarity scores at every level by ignoring the importance of the discriminative feature in the various expert features. Furthermore, it lacks extensive experiments and ablations to justify the contributions. The limitations of \cite{satar_2021} are addressed with the extension in this paper. Our method exploits intra-modality correlation at the global level by preventing noise from local fine-grained local features with self-attention. Likewise, we utilize inter-modality correlations at the local-level features with the Mixture-of-Expert model to combine the local and global video experts. Besides, apply weighted sum to the video experts along with extensive experiments and ablations.

\section{Method}

Figure \ref{fig:model} introduces our model, which has three main parts: textual encoding, video encoding, and text-video matching.

\subsection{Textual Encoding}
To comply with disentangled video features, we also disentangle the textual features hierarchically. Every video clip has at least one descriptive textual sentence in a hierarchically structured way. For instance, while a sentence could define global features, action and entity-related words would define the local features.

By following recent works \cite{satar_2021, Chen_2020_CVPR}, we utilize \cite{shi2019simple} for handling semantic role labeling to disentangle the structure of sentences in the form of graphs. To exploit the semantic relationship among the words, verbs are connected to the sentence by showing the temporal relations with their directed edges, and then the objects are connected to verbs. We define the relationship between actions and entities as $r_{ij}$, in which $i$ refers to action nodes and $j$ refers to verb nodes. Before applying GCN, we use bidirectional LSTM to have contextualized word embeddings. While we apply soft attention for sentence-level embedding, we implement max-pooling over verbs and objects for their corresponding embeddings.  

Rather than using relational GCN \cite{rgcn} by learning weight matrix for each semantic role which would be computationally expensive, we follow Chen \textit{et al.} \cite{Chen_2020_CVPR} to utilize factorised weights in two steps. Firstly, the semantic role matrix $W_r$, which is specific matrix for different semantic roles are multiplied with initialized node embeddings $g_i$ $\epsilon$ \{$g_e$, $g_a$, $g_o$\} in Eq. \ref{eq_text1}. $r_{ij}$ is an one-hot vector referring to the edge type from node $i$ to $j$. $\odot$ refers to element-wise multiplication.

\begin{equation}
g^0_{i} = g_i \odot W_rr_{ij}
\label{eq_text1}
\end{equation}

Secondly, a graph attention network is applied to neighbor nodes. Then $W_t$ matrix, which is shared across all relations types, is used to exploit attended nodes in Eq. \ref{eq_text2}. $\beta$ refers to the outcome after attention is implemented for every node.

\begin{equation}
g^{l+1}_{i} = g^{l}_i + W^{l+1}_t \sum_{j\varepsilon  N_i}\beta_{ij} (g^l_j)
\label{eq_text2}
\end{equation}

After applying these two formulas, we gather the textual representation for global and local features; for sentences, verbs, and words, respectively.

\subsection{Video Encoding}

While disentangling language queries into hierarchical features is relatively easy, parsing videos into hierarchical features could be challenging. We propose different transformer models for global and local levels. While we use only intra-modality features at the global level, we utilize both intra-modality and inter-modality specific features at the local levels. 

The intuition behind this is that while the global features could guide the local representations through the cross-modal attention at the action or object level, the local features may give noise to the global representations when used at the appearance level. We use transformer encoder and decoders to implement this idea as inspired from \cite{NIPS2017_3f5ee243,gabeur2020mmt, devlin-etal-2019-bert, MDVC_Iashin_2020}.

Calculating appearance-level video encodings is relatively easy. The 2D feature $F_S$ is fed into the self-attention layer to encode into $z_s$. Then, a feed-forward layer FF outputs the final contextualized appearance feature. Normalization of the layer is done under the Norm function.

\begin{equation}
\begin{aligned}
&z_s = \textrm{Norm}\big(\textrm{MultiHead}(F_S, F_S, F_S) + F_S\big)& \\
&E_S = \textrm{Norm}\big(\textrm{FF}(z_s) + z_s\big)& \\
\label{decoder}
\end{aligned}  
\end{equation}

We follow Vaswani \textit{et al.} \cite{NIPS2017_3f5ee243}'s multi-headed attention layers as formulated in Eq. \ref{transformer}. During the training process, all the W matrices are learned. While query Q is the same as key K and value V in the self-attention layers, the query Q could be from another source in the cross-modal attention layers. Layer normalization and the residual connection are also applied after every attention layer. 

\begin{equation}
\begin{aligned}
&\textrm{MultiHead}(Q,K,V) = \textrm{Concat}(\textrm{Head}_1, ..., \textrm{Head}_h)W^O \\ 
&\textrm{Head}_i = \textrm{Attention}(QW_i^Q, KW_i^K, VW_i^V) \\ 
&\textrm{Attention}(Q,K,V) = \sigma\Big(\frac{QK^T}{\sqrt{d}}V\Big)
\end{aligned}
\label{transformer}
\end{equation}

Complementary features are applied with cross-modal attention to calculate the verb $E_A$ and noun $E_O$ level encodings. For instance, we explain the formula for verb level in Eq. \ref{eq:encoder}. First, the appearance level $F_S$ and noun level features $F_O$ are concatenated. Then, a multi-headed self-attention is applied, followed by a feed-forward layer FF. This generates context-specific feature $s_e$ of the verb level.

\begin{equation}
\begin{aligned}
&f_e = \textrm{Concat}(F_S,F_O)& \\
&z_e = \textrm{Norm}\big(\textrm{MultiHead}(f_e, f_e, f_e) + f_e\big)& \\ 
&s_e = \textrm{Norm}\big(\textrm{FF}(z_e) + z_e\big)& \\
\label{eq:encoder}
\end{aligned}  
\end{equation}

Then, self-attention of verb level features is calculated as in the first row of Eq. \ref{eq:encoder}.

Finally, cross-modal attention is applied such that the contextualized verb-level features would attend to the contextualized appearance and object-level features. Then, a feed-forward linear layer FF and layer normalization are applied. We repeat the same process to calculate the noun-level embeddings. Then, these three-level video embeddings are used in the video-text matching part.

\begin{equation}
\begin{aligned}
&z_a = \textrm{Norm}\big(\textrm{MultiHead}(F_A, F_A, F_A) + F_A\big)& \\
&c_e = \textrm{Norm}\big(\textrm{MultiHead}(s_e, z_a, z_a) + z_a\big)& \\
&E_A = \textrm{Norm}\big(\textrm{FF}(c_e) + c_e\big)& \\
\label{eq:decoder}
\end{aligned}  
\end{equation}

\subsection{Video-Text Matching}

Some works such as \cite{Chen_2020_CVPR, satar_2021} average the similarities of the different embedding levels, assuming that they are equally important. Other works \cite{gabeur2020mmt, miech18learning} apply weighted sum but predicted from textual encodings. On the other hand, we apply weighted sum to the video encodings as the video encodings, especially at the local level, become contextualized via cross-modal attention with the other modalities. The guidance of the textual encodings may not be necessary. 

We simply use cosine similarity with the video and textual embeddings to calculate the matching score by adding a weighting calculated on video embeddings only. We utilize contrastive ranking loss  \cite{Chen_2020_CVPR} in training by aiming to have positive and negative pairs larger than a pre-set margin. If $v$ and $c$ symbolize visual and textual representation, the positive pairs and the negative pairs could be presented like following, $(v_p,c_p)$ and $(v_p,c_n)$ / $(v_n,c_p)$, respectively. A pre-set margin is referred to as $\Delta$ to calculate the contrastive loss.

\begin{equation}
s(V,C) = \sum^3_{i=1} w_i(V) \frac{<v_i, c_i>}{||v_i||_2 ||c_i||_2 }
\label{weights1}
\end{equation}

\begin{equation}
w_i(V) = \frac{ e^{h(V)^\tau a_i } }{ \sum^3_{j=1} e^{h(V)^\tau a_j } }
\label{weights2}
\end{equation}

$W_i(v)$ represents the weight for the ith video expert feature. We feed the expert encodings into a linear layer and then apply a softmax operation. Since we have three linear layers, we would calculate three different weights. 

\section{Experiments}

We share our experimental results on YouCook2 and MSR-VTT datasets for the video retrieval task and present various ablations results. 

\subsection{Implementation Details}

\textbf{YouCook2 dataset: } This dataset \cite{ZhXuCoAAAI18} is collected from YouTube only from cooking-related videos with 89 different recipes. They are all captured in various kitchen environments from the third-person viewpoint. The video clips are labeled by imperative English sentences along with temporal boundaries referring to the actions. The annotation was done manually by humans. Video clips are extracted by using the formal annotation file. We feed our model with the validation dataset for evaluation following the other studies since the test set is not publicly available. Normally, the dataset has 10,337 clips and 3,492 clips in the training and validation set, respectively. However, some of the raw videos are not available due to being removed or becoming private. This makes ~8\% of training clips and ~7\% of validation clips missing in total. Thus, we use 9,496 clips and 3,266 clips in the training and validation set, respectively.

\textbf{MSR-VTT dataset:} It \cite{xu2016msr-vtt} contains 10K video clips with 20 descriptions for each video and collected by a commercial video search engine. The annotation is done by humans. This dataset has various train/test splits. One of the most used splits is called 1k-B. It has 6,656 clips for training and 1,000 clips for testing. The full split, which is another widely used split, contains 6,513 video clips for training, 497 for validation, and 2990 for testing. While we test our method mostly in this 1k-B split, we also report in the full split as well.

\begin{figure}[!htb]
\centering
\centerline{\includegraphics[width=8cm]{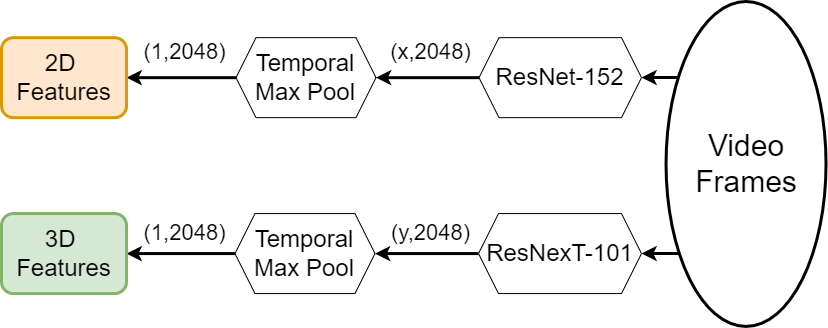}}
\centering
\caption{Feature extraction and pre-processing method for our experiments.}
\label{fig:extraction}
\end{figure}

\textbf{Video features:} We extract our own features from the raw videos of YouCook2 by following Miech \textit{et al.} \cite{miech19howto100m}. To extract appearance features we use two dimensional Convolutional Neural Networks of ResNet-152 \cite{2D_CNNs} model, pre-trained on Imagenet \cite{imagenet}. To extract action level features we utilize three dimensional Convolutional Neural Networks of ResNeXt-101 \cite{3D_CNNs} model, pre-trained on Kinetics \cite{kinetics}. Later on, we implement max-pooling on temporal dimension, making the dimension (1, 2048) for each feature set. The process is also depicted in Figure \ref{fig:extraction}. For MSR-VTT, we gather appearance level features of ResNet-152 model from Chen \textit{et al.} \cite{Chen_2020_CVPR} and action level features of ResNeXt-101 model from Liu \textit{et al.} \cite{Liu2019a}, respectively. 

\textbf{Evaluation metrics.} Given a textual query, we rank all the video candidates such that the video associated with the textual query is ranked as high as possible. Recall (R@k) and Median Rank (MedR) are used to evaluate the model. For instance, R@k is calculated if the ground truth video is in the first kth in the ranking list, where k is \{1,5,10\}. Higher recall values refer to better performance. MedR denotes the median rank of correctly predicted video clips in the list. The lower MedR is, the performance is better.

\textbf{Training.} Glove embeddings \cite{pennington-etal-2014-glove} is used to set the word embedding with the size of 300 for the textual embedding. To parse the text embeddings hierarchically, we follow \cite{Chen_2020_CVPR, satar_2021}. Graph convolutions are applied with two layers, outputting 1024 dimensions. We apply a linear layer to each video expert feature embeddings to be compatible with the textual layer, making their dimension 1024 as well. The $\Delta$ margin is set to $0.2$. Our mini-batch size is 32. While the epoch is 100 for the YouCook2 dataset, it is 50 for the MSR-VTT dataset.

\subsection{Experiments on YouCook2}

Table \ref{tab:yc2-sota} compares our approach with SOTA methods for video retrieval task on YouCook2 dataset. It includes all the methods that report their results without pre-training. The above part of Table \ref{tab:yc2-sota} shows the methods when using the same feature set. We overpass all the recent papers in all metrics, except the last metric for one work \cite{coot}, using the same feature backbone. Please note that the HGR method by \textit{Chen et al.} \cite{Chen_2020_CVPR} is reported by Satar et al. \cite{satar_2021}. All the models in the table use pre-extracted features; in other words, they are not end-to-end models, making the comparison fair.

The lower part of Table \ref{tab:yc2-sota} introduces the results with pre-training and different visual feature backbones. We see higher results for two reasons. UniVL \cite{Luo2020UniVL}, COOT \cite{coot}, and TACo \cite{taco} methods use S3D features pre-trained on HowTo100M. Since HowTo100M is similar to YouCook2 as an instructional video dataset, it would be more useful than pre-training on Kinetics. On the other hand, MMVC \cite{Gabeur2022Masking} uses S3D features pre-trained on Kinetics. However, this model exploits additional audio and Automatic Speech Recognition (ASR) features which could boost the performance.

\begin{table*}[!htb]
\centering
\caption{We Compare Our Approach with SOTA Methods for Video Retrieval on YouCook2 Validation Set. We Overpass Them in All Metrics Except the Last one by Using the Same Feature Set When No Pre-training. The Lower Part Shows the Methods with Pre-training.}
\label{tab:yc2-sota}
\begin{tabular}{|c|c|c|cccc|}
\hline
 &
   &
   &
  \multicolumn{4}{c|}{\textbf{Text-to-Video Retrieval}}  \\ \cline{4-7}
\multirow{-2}{*}{\textbf{Method}} &
  \multirow{-2}{*}{\textbf{Visual Backbone}} &
  \multirow{-2}{*}{\textbf{\begin{tabular}[c]{@{}c@{}}Batch Size\end{tabular}}} &
  \multicolumn{1}{c|}{\textbf{R@1↑}} &
  \multicolumn{1}{c|}{\textbf{R@5↑}} &
  \multicolumn{1}{c|}{\textbf{R@10↑}} &
  \multicolumn{1}{c|}{\textbf{MdR↑}} \\ \hline \hline
\rowcolor[HTML]{FFFFFF} 
Random &
  - &
  - &
  \multicolumn{1}{c|}{\cellcolor[HTML]{FFFFFF}0.03} &
  \multicolumn{1}{c|}{\cellcolor[HTML]{FFFFFF}0.15} &
  \multicolumn{1}{c|}{\cellcolor[HTML]{FFFFFF}0.3} &
  1675 \\ \hline
\rowcolor[HTML]{FFFFFF} 
\begin{tabular}[c]{@{}c@{}}UniVL (v2-FT-Joint), 2020 \cite{Luo2020UniVL}  \end{tabular} &
  \begin{tabular}[c]{@{}c@{}}Resnet-152 (ImageNet) + \\ ResNeXt-101 (Kinetics)\end{tabular} &
  32 &
  \multicolumn{1}{c|}{\cellcolor[HTML]{FFFFFF}3.4} &
  \multicolumn{1}{c|}{\cellcolor[HTML]{FFFFFF}10.8} &
  \multicolumn{1}{c|}{\cellcolor[HTML]{FFFFFF}17.8} &
  76 \\ \hline
\rowcolor[HTML]{FFFFFF} 
Satar et al., 2021 \cite{satar_2021} &
  \begin{tabular}[c]{@{}c@{}}Resnet-152 (ImageNet) + \\ ResNeXt-101 (Kinetics)\end{tabular} &
  32 &
  \multicolumn{1}{c|}{\cellcolor[HTML]{FFFFFF}4.5} &
  \multicolumn{1}{c|}{\cellcolor[HTML]{FFFFFF}13.2} &
  \multicolumn{1}{c|}{\cellcolor[HTML]{FFFFFF}20.0} &
  85 \\ \hline
\rowcolor[HTML]{FFFFFF} 
Miech et al., 2019 \cite{miech19howto100m} &
  \begin{tabular}[c]{@{}c@{}}Resnet-152 (ImageNet) + \\ ResNeXt-101 (Kinetics)\end{tabular} &
  - &
  \multicolumn{1}{c|}{\cellcolor[HTML]{FFFFFF}4.2} &
  \multicolumn{1}{c|}{\cellcolor[HTML]{FFFFFF}13.7} &
  \multicolumn{1}{c|}{\cellcolor[HTML]{FFFFFF}21.5} &
  65 \\ \hline
\rowcolor[HTML]{FFFFFF} 
HGR*, 2020 \cite{Chen_2020_CVPR} &
  \begin{tabular}[c]{@{}c@{}}Resnet-152 (ImageNet) + \\ ResNeXt-101 (Kinetics)\end{tabular} &
  32 &
  \multicolumn{1}{c|}{\cellcolor[HTML]{FFFFFF}4.8} &
  \multicolumn{1}{c|}{\cellcolor[HTML]{FFFFFF}14.0} &
  \multicolumn{1}{c|}{\cellcolor[HTML]{FFFFFF}20.3} &
  85 \\ \hline
\rowcolor[HTML]{FFFFFF} 
HGR*, 2020 \cite{Chen_2020_CVPR} &
  \begin{tabular}[c]{@{}c@{}}Resnet-152 (ImageNet) + \\ ResNeXt-101 (Kinetics) + \\ Faster R-CNN (MS COCO)\end{tabular} &
  32 &
  \multicolumn{1}{c|}{\cellcolor[HTML]{FFFFFF}4.7} &
  \multicolumn{1}{c|}{\cellcolor[HTML]{FFFFFF}14.1} &
  \multicolumn{1}{c|}{\cellcolor[HTML]{FFFFFF}20.0} &
  87 \\ \hline
\rowcolor[HTML]{FFFFFF} 
HGLMM, 2015 \cite{hglmm} &
  Fisher Vectors &
  - &
  \multicolumn{1}{c|}{\cellcolor[HTML]{FFFFFF}4.6} &
  \multicolumn{1}{c|}{\cellcolor[HTML]{FFFFFF}14.3} &
  \multicolumn{1}{c|}{\cellcolor[HTML]{FFFFFF}21.6} &
  75 \\ \hline
\rowcolor[HTML]{FFFFFF} 
Satar et al., 2021 \cite{satar_2021} &
  \begin{tabular}[c]{@{}c@{}}Resnet-152 (ImageNet) + \\ ResNeXt-101 (Kinetics) + \\ Faster R-CNN (MS COCO)\end{tabular} &
  32 &
  \multicolumn{1}{c|}{\cellcolor[HTML]{FFFFFF}5.3} &
  \multicolumn{1}{c|}{\cellcolor[HTML]{FFFFFF}14.5} &
  \multicolumn{1}{c|}{\cellcolor[HTML]{FFFFFF}20.8} &
  77 \\ \hline
\rowcolor[HTML]{FFFFFF} 
SwAMP, 2021 \cite{swamp} &
  \begin{tabular}[c]{@{}c@{}}Resnet-152 (ImageNet) + \\ ResNeXt-101 (Kinetics)\end{tabular} &
  128 &
  \multicolumn{1}{c|}{\cellcolor[HTML]{FFFFFF}4.8} &
  \multicolumn{1}{c|}{\cellcolor[HTML]{FFFFFF}14.5} &
  \multicolumn{1}{c|}{\cellcolor[HTML]{FFFFFF}22.5} &
  57 \\ \hline
\rowcolor[HTML]{FFFFFF} 
TACo, 2021 \cite{taco} &
  \begin{tabular}[c]{@{}c@{}}Resnet-152 (ImageNet) + \\ ResNeXt-101 (Kinetics)\end{tabular} &
  128 &
  \multicolumn{1}{c|}{\cellcolor[HTML]{FFFFFF}4.9} &
  \multicolumn{1}{c|}{\cellcolor[HTML]{FFFFFF}14.7} &
  \multicolumn{1}{c|}{\cellcolor[HTML]{FFFFFF}21.7} &
  63 \\ \hline
\rowcolor[HTML]{FFFFFF} 
COOT, 2020 \cite{coot} &
  \begin{tabular}[c]{@{}c@{}}Resnet-152 (ImageNet) + \\ ResNeXt-101 (Kinetics)\end{tabular} &
  32 &
  \multicolumn{1}{c|}{\cellcolor[HTML]{FFFFFF}5.9} &
  \multicolumn{1}{c|}{\cellcolor[HTML]{FFFFFF}16.7} &
  \multicolumn{1}{c|}{\cellcolor[HTML]{FFFFFF}24.8} &
  \textbf{49.7} \\ \hline
\rowcolor[HTML]{FFFFFF} 
\textbf{RoME} &
  \begin{tabular}[c]{@{}c@{}}Resnet-152 (ImageNet) + \\ ResNeXt-101 (Kinetics)\end{tabular} &
  32 &
  \multicolumn{1}{c|}{\cellcolor[HTML]{FFFFFF}\textbf{6.3}} &
  \multicolumn{1}{c|}{\cellcolor[HTML]{FFFFFF}\textbf{16.9}} &
  \multicolumn{1}{c|}{\cellcolor[HTML]{FFFFFF}\textbf{25.2}} &
  53 \\ \hline \hline
\rowcolor[HTML]{EFEFEF} 
UniVL (FT-Joint), 2020 \cite{Luo2020UniVL}  &
  S3D (HowTo100M) &
  32 &
  \multicolumn{1}{c|}{\cellcolor[HTML]{EFEFEF}7.7} &
  \multicolumn{1}{c|}{\cellcolor[HTML]{EFEFEF}23.9} &
  \multicolumn{1}{c|}{\cellcolor[HTML]{EFEFEF}34.7} &
  21 \\ \hline
\rowcolor[HTML]{EFEFEF} 
MMCV, 2022 \cite{Gabeur2022Masking} &
  S3D (Kinetics) &
  32 &
  \multicolumn{1}{c|}{\cellcolor[HTML]{EFEFEF}16.6} &
  \multicolumn{1}{c|}{\cellcolor[HTML]{EFEFEF}37.4} &
  \multicolumn{1}{c|}{\cellcolor[HTML]{EFEFEF}48.3} &
  12 \\ \hline
\rowcolor[HTML]{EFEFEF} 
COOT, 2020 \cite{coot} &
  S3D (HowTo100M) &
  64 &
  \multicolumn{1}{c|}{\cellcolor[HTML]{EFEFEF}16.7} &
  \multicolumn{1}{c|}{\cellcolor[HTML]{EFEFEF}40.2} &
  \multicolumn{1}{c|}{\cellcolor[HTML]{EFEFEF}52.3} &
  9 \\ \hline
\rowcolor[HTML]{EFEFEF} 
TACo, 2021 \cite{taco} &
  S3D (HowTo100M) &
  128 &
  \multicolumn{1}{c|}{\cellcolor[HTML]{EFEFEF}16.6} &
  \multicolumn{1}{c|}{\cellcolor[HTML]{EFEFEF}40.3} &
  \multicolumn{1}{c|}{\cellcolor[HTML]{EFEFEF}53.1} &
  9 \\ \hline
\end{tabular}%
\end{table*}

\subsection{Experiments on MSR-VTT}

Table \ref{tab:msrvtt_sota} compares our approach with SOTA methods when using the same feature set on the MSR-VTT dataset. These include all the methods that report text-to-video retrieval results without pre-training. We overpass all the methods in all metrics. In particular, our method surpasses the performance by 2.9\% on R@1 over the CE method \cite{Liu2019a}, which uses four features, even though we only use two of them. Moreover, our method improves the performance by 1.9\% on R@1 over the best SOTA approach \cite{taco}.

For the full split, we overpass all the methods that we have similar feature sets. However, we surpass the CE method \cite{Liu2019a} only in the first two metrics. One of the main reasons that CE overpasses our method in the last two metrics could be the advantage of using various other visual experts such as audio and ASR. Secondly, the testing set becomes tripled than 1k-B split, increasing the advantage of having many visual expert features. Thirdly, our models' textual encoding part may not handle having twenty captions for each video clip, as it is included in MSR-VTT. A transformer-based approach such as BERT may perform better, although we did not apply any experiments with BERT. Moreover, it is hard to make a comparison among some approaches. For example, Alayrac \textit{et al.} \cite{miech20endtoend} report their result when pre-trained on HowTo100M and a zero-shot setting. TeachText-CE \cite{croitoru2021teachtext} uses various textual feature experts as long as seven different visual feature experts, making it hard to compare with our result, which is still added to the table for reference. Nevertheless, we overpass the CE method in the first two metrics, which could be more critical in real-world applications.

\begin{table*}[!htb]
\centering
\caption{We Compare Our Approach with SOTA Methods for Video Retrieval on MSR-VTT 1k-B Split and Full Split. We Overpass them in All Metrics When No Pre-training With the Same Feature Set on 1k-B Split. On Full Split, We only Overpass them in the First Two Metrics, which Are More Functional to Real-world Applications.}
\label{tab:msrvtt_sota}
\begin{tabular}{|ccccccc|}
\hline
\multicolumn{1}{|c|}{}                                  & \multicolumn{1}{c|}{}                                                                                                              & \multicolumn{1}{c|}{}                                      & \multicolumn{4}{c|}{\textbf{Text-to-Video Retrieval}}                                                                         \\ \cline{4-7} 
\multicolumn{1}{|c|}{\multirow{-2}{*}{\textbf{Method}}} & \multicolumn{1}{c|}{\multirow{-2}{*}{\textbf{Visual Backbone}}}                                                                    & \multicolumn{1}{c|}{\multirow{-2}{*}{\textbf{Batch Size}}} & \multicolumn{1}{c|}{\textbf{R@1↑}} & \multicolumn{1}{c|}{\textbf{R@5↑}} & \multicolumn{1}{c|}{\textbf{R@10↑}} & \textbf{MdR↓} \\ \hline \hline
\multicolumn{7}{|c|}{\cellcolor[HTML]{EFEFEF}\textbf{Split 1k-B}}    \\ \hline
\multicolumn{1}{|c|}{Random}                            & \multicolumn{1}{c|}{-}                                                                                                             & \multicolumn{1}{c|}{-}                                     & \multicolumn{1}{c|}{0.1}           & \multicolumn{1}{c|}{0.5}           & \multicolumn{1}{c|}{1.0}            & 500           \\ \hline
\multicolumn{1}{|c|}{MoEE, 2018 \cite{miech18learning}}                        & \multicolumn{1}{c|}{\begin{tabular}[c]{@{}c@{}}SENet-154 (ImageNet) +\\ R(2+1)D (IG-65m) +\end{tabular}}                           & \multicolumn{1}{c|}{-}                                     & \multicolumn{1}{c|}{13.6}          & \multicolumn{1}{c|}{37.9}          & \multicolumn{1}{c|}{51.0}           & 10            \\ \hline
\multicolumn{1}{|c|}{JPoSE, 2019 \cite{wray2019fine}}                       & \multicolumn{1}{c|}{TSN + Flow}                                                                                                    & \multicolumn{1}{c|}{-}                                     & \multicolumn{1}{c|}{14.3}          & \multicolumn{1}{c|}{38.1}          & \multicolumn{1}{c|}{53.0}           & 9             \\ \hline
\multicolumn{1}{|c|}{UniVL (v1*), 2020 \cite{Luo2020UniVL}}                 & \multicolumn{1}{c|}{\begin{tabular}[c]{@{}c@{}}Resnet-152 (ImageNet) + \\ ResNeXt-101 (Kinetics)\end{tabular}}                     & \multicolumn{1}{c|}{-}                                     & \multicolumn{1}{c|}{14.6}          & \multicolumn{1}{c|}{39.0}          & \multicolumn{1}{c|}{52.6}           & 10            \\ \hline
\multicolumn{1}{|c|}{SwAMP, 2021 \cite{swamp}}                       & \multicolumn{1}{c|}{\begin{tabular}[c]{@{}c@{}}Resnet-152 (ImageNet) + \\ ResNeXt-101 (Kinetics)\end{tabular}}                     & \multicolumn{1}{c|}{128}                                   & \multicolumn{1}{c|}{15.0}          & \multicolumn{1}{c|}{38.5}          & \multicolumn{1}{c|}{50.3}           & 10            \\ \hline
\multicolumn{1}{|c|}{CE, 2019 \cite{Liu2019a}}                          & \multicolumn{1}{c|}{\begin{tabular}[c]{@{}c@{}}SENet-154 (ImageNet) +\\ R(2+1)D (IG-65m) +\\ Six more visual experts\end{tabular}} & \multicolumn{1}{c|}{-}                                     & \multicolumn{1}{c|}{18.2}          & \multicolumn{1}{c|}{46.0}          & \multicolumn{1}{c|}{60.7}           & 7             \\ \hline
\multicolumn{1}{|c|}{TACo, 2021 \cite{taco}}                        & \multicolumn{1}{c|}{\begin{tabular}[c]{@{}c@{}}Resnet-152 (ImageNet) + \\ ResNeXt-101 (Kinetics)\end{tabular}}                     & \multicolumn{1}{c|}{128}                                   & \multicolumn{1}{c|}{19.2}          & \multicolumn{1}{c|}{44.7}          & \multicolumn{1}{c|}{57.2}           & 7             \\ \hline
\multicolumn{1}{|c|}{\textbf{RoME}}                     & \multicolumn{1}{c|}{\begin{tabular}[c]{@{}c@{}}Resnet-152 (ImageNet) + \\ ResNeXt-101 (Kinetics)\end{tabular}}                     & \multicolumn{1}{c|}{\textbf{32}}                           & \multicolumn{1}{c|}{\textbf{21.1}} & \multicolumn{1}{c|}{\textbf{50.0}} & \multicolumn{1}{c|}{\textbf{63.1}}  & \textbf{5}    \\ \hline \hline
\multicolumn{7}{|c|}{\cellcolor[HTML]{EFEFEF}\textbf{Full Split}}   \\ \hline
\multicolumn{1}{|c|}{Random}                            & \multicolumn{1}{c|}{-}                                                                                                             & \multicolumn{1}{c|}{-}                                     & \multicolumn{1}{c|}{0.1}           & \multicolumn{1}{c|}{0.5}           & \multicolumn{1}{c|}{1.0}            & 500           \\ \hline
\multicolumn{1}{|c|}{Dual Encoding, 2019 \cite{dong_cvpr19}}               & \multicolumn{1}{c|}{Resnet-152 (ImageNet)}                                                                                         & \multicolumn{1}{c|}{128}                                   & \multicolumn{1}{c|}{7.7}           & \multicolumn{1}{c|}{22.0}          & \multicolumn{1}{c|}{31.8}           & 32.0          \\ \hline
\multicolumn{1}{|c|}{Alayrac \textit{et al.}, 2020 \cite{miech20endtoend}}              & \multicolumn{1}{c|}{\begin{tabular}[c]{@{}c@{}}Resnet-152 (ImageNet) + \\ ResNeXt-101 (Kinetics)\end{tabular}}                     & \multicolumn{1}{c|}{-}                                     & \multicolumn{1}{c|}{9.9}           & \multicolumn{1}{c|}{24.0}          & \multicolumn{1}{c|}{32.4}           & 29.5          \\ \hline
\multicolumn{1}{|c|}{HGR, 2020 \cite{Chen_2020_CVPR}}                         & \multicolumn{1}{c|}{Resnet-152 (ImageNet)}                                                                                         & \multicolumn{1}{c|}{128}                                   & \multicolumn{1}{c|}{9.2}           & \multicolumn{1}{c|}{26.2}          & \multicolumn{1}{c|}{36.5}           & 24            \\ \hline
\multicolumn{1}{|c|}{CE, 2019 \cite{Liu2019a}}                          & \multicolumn{1}{c|}{\begin{tabular}[c]{@{}c@{}}SENet-154 (ImageNet) +\\ R(2+1)D (IG-65m) +\\ Six more visual experts\end{tabular}} & \multicolumn{1}{c|}{-}                                     & \multicolumn{1}{c|}{10.0}          & \multicolumn{1}{c|}{29.0}          & \multicolumn{1}{c|}{\textbf{42.2}}  & \textbf{16}   \\ \hline
\multicolumn{1}{|c|}{\textbf{RoME}}                     & \multicolumn{1}{c|}{\begin{tabular}[c]{@{}c@{}}Resnet-152 (ImageNet) + \\ ResNeXt-101 (Kinetics)\end{tabular}}                     & \multicolumn{1}{c|}{32}                                    & \multicolumn{1}{c|}{\textbf{10.7}} & \multicolumn{1}{c|}{\textbf{29.6}} & \multicolumn{1}{c|}{41.2}           & 17            \\ \hline \hline
\multicolumn{1}{|c|}{TeachText-CE, 2021 \cite{croitoru2021teachtext}}                & \multicolumn{1}{c|}{\begin{tabular}[c]{@{}c@{}}SENet-154 (ImageNet) +\\ R(2+1)D (IG-65m) +\\ Six more visual experts\end{tabular}} & \multicolumn{1}{c|}{64}                                    & \multicolumn{1}{c|}{11.8}          & \multicolumn{1}{c|}{32.7}          & \multicolumn{1}{c|}{45.3}           & 13.0          \\ \hline
\end{tabular}
\end{table*}

\subsection{Experiments for Video-to-Text Retrieval}

Although the video-to-text retrieval task is not our primary motivation, we report it in Table \ref{tab:yc2_video_to_text} to show that our model could also be used in this aim. For YouCook2, only one method reports their video-to-text retrieval results when there is no pre-training, which is reported by \cite{satar_2021}. We overpass the compared methods in all metrics with a high margin on both datasets without pre-training. 

\begin{table*}[!htb]
\centering
\caption{We Compare Our Approach with SOTA Methods for Video-to-Text Retrieval on YouCook2 Validation Set and MSR-VTT 1k-B Split. We Overpass them in all Metrics When there is no Pre-training with the Same Feature Set.}
\label{tab:yc2_video_to_text}
\begin{tabular}{|ccccccc|}
\hline
\multicolumn{1}{|c|}{}                                  & \multicolumn{1}{c|}{}                                                                                                                      & \multicolumn{1}{c|}{}                                      & \multicolumn{4}{c|}{\textbf{Video-to-Text Retrieval}}                                                                         \\ \cline{4-7} 
\multicolumn{1}{|c|}{\multirow{-2}{*}{\textbf{Method}}} & \multicolumn{1}{c|}{\multirow{-2}{*}{\textbf{Visual Backbone}}}                                                                            & \multicolumn{1}{c|}{\multirow{-2}{*}{\textbf{Batch Size}}} & \multicolumn{1}{c|}{\textbf{R@1↑}} & \multicolumn{1}{c|}{\textbf{R@5↑}} & \multicolumn{1}{c|}{\textbf{R@10↑}} & \textbf{MdR↓} \\ \hline \hline
\multicolumn{7}{|c|}{\cellcolor[HTML]{EFEFEF}\textbf{YouCook2}}                                                                                                                                                                                                                                                                                                                    \\ \hline
\multicolumn{1}{|c|}{HGR*, 2020 \cite{Chen_2020_CVPR}}                              & \multicolumn{1}{c|}{\begin{tabular}[c]{@{}c@{}}Resnet-152 (ImageNet) + \\ ResNeXt-101 (Kinetics)\end{tabular}}                             & \multicolumn{1}{c|}{32}                                    & \multicolumn{1}{c|}{4.1}           & \multicolumn{1}{c|}{13.0}          & \multicolumn{1}{c|}{19.0}           & 85            \\ \hline
\multicolumn{1}{|c|}{HGR*, 2020 \cite{Chen_2020_CVPR}}                              & \multicolumn{1}{c|}{\begin{tabular}[c]{@{}c@{}}Resnet-152 (ImageNet) + \\ ResNeXt-101 (Kinetics) + \\ Faster R-CNN (MS COCO)\end{tabular}} & \multicolumn{1}{c|}{32}                                    & \multicolumn{1}{c|}{4.2}           & \multicolumn{1}{c|}{13.2}          & \multicolumn{1}{c|}{18.9}           & 84            \\ \hline
\multicolumn{1}{|c|}{\textbf{RoME}}                     & \multicolumn{1}{c|}{\begin{tabular}[c]{@{}c@{}}Resnet-152 (ImageNet) + \\ ResNeXt-101 (Kinetics)\end{tabular}}                             & \multicolumn{1}{c|}{32}                                    & \multicolumn{1}{c|}{\textbf{4.9}}  & \multicolumn{1}{c|}{\textbf{15.9}} & \multicolumn{1}{c|}{\textbf{24.1}}  & \textbf{55}   \\ \hline \hline
\multicolumn{7}{|c|}{\cellcolor[HTML]{EFEFEF}\textbf{MSR-VTT}}                                                           \\ \hline 
\multicolumn{1}{|c|}{JPoSE, 2019 \cite{wray2019fine}}                             & \multicolumn{1}{c|}{TSN + Flow}                                                                                                            & \multicolumn{1}{c|}{-}                                     & \multicolumn{1}{c|}{16.4}          & \multicolumn{1}{c|}{41.3}          & \multicolumn{1}{c|}{54.4}           & 8.7           \\ \hline
\multicolumn{1}{|c|}{CE, 2019 \cite{Liu2019a}}                                & \multicolumn{1}{c|}{\begin{tabular}[c]{@{}c@{}}SENet-154 (ImageNet) +\\ R(2+1)D (IG-65m) +\\ Six more visual experts\end{tabular}}         & \multicolumn{1}{c|}{-}                                     & \multicolumn{1}{c|}{18.0}          & \multicolumn{1}{c|}{46.0}          & \multicolumn{1}{c|}{60.3}           & 6.5           \\ \hline
\multicolumn{1}{|c|}{\textbf{RoME}}                     & \multicolumn{1}{c|}{\begin{tabular}[c]{@{}c@{}}Resnet-152 (ImageNet) + \\ ResNeXt-101 (Kinetics)\end{tabular}}                             & \multicolumn{1}{c|}{32}                                    & \multicolumn{1}{c|}{\textbf{27.9}} & \multicolumn{1}{c|}{\textbf{58.5}} & \multicolumn{1}{c|}{\textbf{73.1}}  & \textbf{4}    \\ \hline
\end{tabular}
\end{table*}

\subsection{Ablation Studies}

We share extensive ablation studies about design choices on weighted sum, model design, and video feature settings.

\subsubsection{Ablation on Weighting Encodings}

Some recent papers add weighting to only textual representation. We do an ablation study by adding weighting to the only visual representation, only textual representation, and both. Our results in Table \ref{tab:abl-weight} indicate: 1) Applying weighted sum over textual or visual encodings generally performs better than average weighting. It justifies the idea that every embedding level impacts the model differently. 2) Weighted sum on only visual encodings brings slightly higher results than having weighted sum on only textual encodings. We assume it is because it ignores discriminate signals from the video experts by making the model text-dependent when using weighted sum on only textual encodings.

\begin{table*}[!htb]
\centering
\caption{Ablation for Weighting Options for Visual and Textual Encodings on YouCook2 Validation Set and MSR-VTT 1k-B Split. 'Average Weighting' Option Means that We Average Both Visual and Textual Encodings After Calculating the Similarity Score. 'Weighted Sum on Only Visual Encodings' Means that We Apply Weighted Sum on Visual Encodings While Averaging the Textual Encodings Before Calculating the Similarity Score. Implementing Weighted Sum on Visual Encodings Boost the Performance.}
\label{tab:abl-weight}
\begin{tabular}{|cccccccc|}
\hline
\multicolumn{1}{|c|}{}                                                                                             & \multicolumn{3}{c|}{\textbf{Visual  Features}}                                                                         & \multicolumn{1}{c|}{}                                & \multicolumn{1}{c|}{}                                & \multicolumn{1}{c|}{}                                 &                                 \\ \cline{2-4}
\multicolumn{1}{|c|}{\multirow{-2}{*}{\textbf{Weighting Option}}}                                                  & \multicolumn{1}{c|}{\textbf{Appearance}} & \multicolumn{1}{c|}{\textbf{Action}} & \multicolumn{1}{c|}{\textbf{Object}} & \multicolumn{1}{c|}{\multirow{-2}{*}{\textbf{R@1↑}}} & \multicolumn{1}{c|}{\multirow{-2}{*}{\textbf{R@5↑}}} & \multicolumn{1}{c|}{\multirow{-2}{*}{\textbf{R@10↑}}} & \multirow{-2}{*}{\textbf{MdR↓}} \\ \hline \hline
\multicolumn{8}{|c|}{\cellcolor[HTML]{EFEFEF}\textbf{YouCook2}}                                                                                                                                                                                                                                                                                                                                                                                     \\ \hline
\multicolumn{1}{|c|}{Average}                                                                                      & \multicolumn{1}{c|}{2D + 3D}             & \multicolumn{1}{c|}{2D + 3D}         & \multicolumn{1}{c|}{2D + 3D}         & \multicolumn{1}{c|}{6}                               & \multicolumn{1}{c|}{16.7}                            & \multicolumn{1}{c|}{24.6}                             & 55                              \\ \hline
\multicolumn{1}{|c|}{\begin{tabular}[c]{@{}c@{}}Weighted Sum on both \\ textual and visual encodings\end{tabular}} & \multicolumn{1}{c|}{2D + 3D}             & \multicolumn{1}{c|}{2D + 3D}         & \multicolumn{1}{c|}{2D + 3D}         & \multicolumn{1}{c|}{5.9}                             & \multicolumn{1}{c|}{16.8}                            & \multicolumn{1}{c|}{24.7}                             & 54                              \\ \hline
\multicolumn{1}{|c|}{\begin{tabular}[c]{@{}c@{}}Weighted Sum on \\ only textual encodings\end{tabular}}            & \multicolumn{1}{c|}{2D + 3D}             & \multicolumn{1}{c|}{2D + 3D}         & \multicolumn{1}{c|}{2D + 3D}         & \multicolumn{1}{c|}{6.1}                             & \multicolumn{1}{c|}{\textbf{17}}                     & \multicolumn{1}{c|}{24.8}                             & \textbf{52}                     \\ \hline
\multicolumn{1}{|c|}{\begin{tabular}[c]{@{}c@{}}Weighted Sum on \\ only visual encodings\end{tabular}}             & \multicolumn{1}{c|}{2D + 3D}             & \multicolumn{1}{c|}{2D + 3D}         & \multicolumn{1}{c|}{2D + 3D}         & \multicolumn{1}{c|}{\textbf{6.3}}                    & \multicolumn{1}{c|}{16.9}                            & \multicolumn{1}{c|}{\textbf{25.2}}                    & 53                              \\ \hline \hline
\multicolumn{8}{|c|}{\cellcolor[HTML]{EFEFEF}\textbf{MSR-VTT}}                                                                                                                                                                                                                                                                                                                                                                                      \\ \hline
\multicolumn{1}{|c|}{Average}                                                                                      & \multicolumn{1}{c|}{2D + 3D}             & \multicolumn{1}{c|}{2D + 3D}         & \multicolumn{1}{c|}{2D + 3D}         & \multicolumn{1}{c|}{20.8}                            & \multicolumn{1}{c|}{49.1}                            & \multicolumn{1}{c|}{62.8}                             & 6                               \\ \hline
\multicolumn{1}{|c|}{\begin{tabular}[c]{@{}c@{}}Weighted Sum on both \\ textual and visual encodings\end{tabular}} & \multicolumn{1}{c|}{2D + 3D}             & \multicolumn{1}{c|}{2D + 3D}         & \multicolumn{1}{c|}{2D + 3D}         & \multicolumn{1}{c|}{20.6}                            & \multicolumn{1}{c|}{49.0}                            & \multicolumn{1}{c|}{62.3}                             & 6                               \\ \hline
\multicolumn{1}{|c|}{\begin{tabular}[c]{@{}c@{}}Weighted Sum on \\ only textual encodings\end{tabular}}            & \multicolumn{1}{c|}{2D + 3D}             & \multicolumn{1}{c|}{2D + 3D}         & \multicolumn{1}{c|}{2D + 3D}         & \multicolumn{1}{c|}{21.1}                            & \multicolumn{1}{c|}{49.7}                            & \multicolumn{1}{c|}{62.9}                             & 6                               \\ \hline
\multicolumn{1}{|c|}{\begin{tabular}[c]{@{}c@{}}Weighted Sum on \\ only visual encodings\end{tabular}}             & \multicolumn{1}{c|}{2D + 3D}             & \multicolumn{1}{c|}{2D + 3D}         & \multicolumn{1}{c|}{2D + 3D}         & \multicolumn{1}{c|}{\textbf{21.1}}                   & \multicolumn{1}{c|}{\textbf{50.0}}                   & \multicolumn{1}{c|}{\textbf{63.1}}                    & \textbf{5}                      \\ \hline
\end{tabular}
\end{table*}

If we examine the calculated weights of each visual level on the YouCook2 dataset, the results are around 0.45, 0.30, and 0.25  for appearance, action, and object levels, respectively. The weights are calculated in Eq. \ref{weights1} and \ref{weights2}, Sec.III. While the 2D feature contributes the most, the other two feature levels affect the results considerably and similarly. It is because the dataset contains crucial objects and actions. The calculated weights of the visual encodings in the MSR-VTT dataset are around 0.47, 0.43, and 0.10 for appearance action and object levels, showing the same importance for appearance features. However, while action features contribute more, the contribution of object-level features decreases sharply. It is mainly due to the nature of the dataset, where there are fewer objects but more continual actions. 

\subsubsection{Ablation on Model Design}

Our proposed model suggests that different attention methods for global and local feature levels increase the result. We could see this increase in Table \ref{tab:abl-model-design}. The model we use for ablation studies applies only self-attention for every level. We compare this model with our proposed model, where we use self-attention in the appearance level but cross-modal and self-attention for the other two levels. Average weighting is applied for both models. All the feature experts have the dimensions of 2048, which makes the comparison fair. The results are improved by proposing different attention mechanisms for global and local levels.

\begin{table*}[!htb]
\centering
\caption{Ablation Study for Model Design Choice on YouCook2 Validation Set and MSR-VTT 1k-B Split. Suggesting different attention mechanisms for global and local levels improves the result.}
\label{tab:abl-model-design}
\begin{tabular}{|cccccccc|}
\hline
\multicolumn{1}{|c|}{}                                                                                                                          & \multicolumn{3}{c|}{\textbf{Visual  Features}}                                                                         & \multicolumn{1}{c|}{}                                & \multicolumn{1}{c|}{}                                & \multicolumn{1}{c|}{}                                 &                                 \\ \cline{2-4}
\multicolumn{1}{|c|}{\multirow{-2}{*}{\textbf{Attention}}}                                                                                      & \multicolumn{1}{c|}{\textbf{Appearance}} & \multicolumn{1}{c|}{\textbf{Action}} & \multicolumn{1}{c|}{\textbf{Object}} & \multicolumn{1}{c|}{\multirow{-2}{*}{\textbf{R@1↑}}} & \multicolumn{1}{c|}{\multirow{-2}{*}{\textbf{R@5↑}}} & \multicolumn{1}{c|}{\multirow{-2}{*}{\textbf{R@10↑}}} & \multirow{-2}{*}{\textbf{MdR↓}} \\ \hline \hline
\multicolumn{8}{|c|}{\cellcolor[HTML]{EFEFEF}\textbf{YouCook2}}                                                                                       \\ \hline
\multicolumn{1}{|c|}{Self-attention for all levels}                                                                                             & \multicolumn{1}{c|}{2D + 3D}             & \multicolumn{1}{c|}{2D + 3D}         & \multicolumn{1}{c|}{2D + 3D}         & \multicolumn{1}{c|}{6}                               & \multicolumn{1}{c|}{16.2}                            & \multicolumn{1}{c|}{24.5}                             & 55                              \\ \hline
\multicolumn{1}{|c|}{\begin{tabular}[c]{@{}c@{}}Self-attention in the appearance level,\\ Cross-modal + Self-attention for others\end{tabular}} & \multicolumn{1}{c|}{2D + 3D}             & \multicolumn{1}{c|}{2D + 3D}         & \multicolumn{1}{c|}{2D + 3D}         & \multicolumn{1}{c|}{\textbf{6}}                      & \multicolumn{1}{c|}{\textbf{16.7}}                   & \multicolumn{1}{c|}{\textbf{24.6}}                    & \textbf{55}                     \\ \hline \hline
\multicolumn{8}{|c|}{\cellcolor[HTML]{EFEFEF}\textbf{MSR-VTT}}        \\ \hline
\multicolumn{1}{|c|}{Self-attention for all levels}                                                                                             & \multicolumn{1}{c|}{2D + 3D}             & \multicolumn{1}{c|}{2D + 3D}         & \multicolumn{1}{c|}{2D + 3D}         & \multicolumn{1}{c|}{20.4}                            & \multicolumn{1}{c|}{48.1}                            & \multicolumn{1}{c|}{62.6}                             & \multicolumn{1}{c|}{6}          \\ \hline
\multicolumn{1}{|c|}{\begin{tabular}[c]{@{}c@{}}Self-attention in the appearance level,\\ Cross-modal + Self-attention for others\end{tabular}} & \multicolumn{1}{c|}{2D + 3D}             & \multicolumn{1}{c|}{2D + 3D}         & \multicolumn{1}{c|}{2D + 3D}         & \multicolumn{1}{c|}{\textbf{20.8}}                   & \multicolumn{1}{c|}{\textbf{49.1}}                   & \multicolumn{1}{c|}{\textbf{62.8}}                    & \multicolumn{1}{c|}{\textbf{6}} \\ \hline
\end{tabular}
\end{table*}

\subsubsection{Ablation on Visual Feature Settings}

While all the variations in our model surpass the baseline in the same feature set, the highest result comes with a concatenated 2D and 3D feature setting in the Table \ref{tab:yc2_ablation_feature}, showing that early fusion gives a boost since the features are correlated. Average weighting is applied to the models.

\begin{table*}[!htb]
\centering
\caption{Ablation on Visual Feature Setting on YouCook2 Dataset. Concatenated Appearance and Action Features Brings an Increase, Rather Than Using Them in the Different Levels Separately.}
\label{tab:yc2_ablation_feature}
\begin{tabular}{|
>{}c |
>{}c 
>{}c 
>{}c |
>{}c |
>{}c |
>{}c |
>{}c |}
\hline
   &
   &
  \multicolumn{1}{c}{\textbf{Visual  Features}}  &
   &
   &
   &
   &
   \\ \cline{2-4}
  \multirow{-2}{*}{\textbf{Attention}} &
  \multicolumn{1}{c|}{\textbf{Appearance}} &
  \multicolumn{1}{c|}{\textbf{Action}} &
  \textbf{Object} &
  \multirow{-2}{*}{\textbf{R@1↑}} &
  \multirow{-2}{*}{\textbf{R@5↑}} &
  \multirow{-2}{*}{\textbf{R@10↑}} &
  \multirow{-2}{*}{\textbf{MdR↓}} \\ \hline \hline
  \begin{tabular}[c]{@{}c@{}}Self-attention in the appearance level,\\ Cross-modal + Self-attention for others\end{tabular} &
  \multicolumn{1}{c|}{2D} &
  \multicolumn{1}{c|}{2D} &
  2D &
  5.7 &
  16.7 &
  24.1 &
  56 \\ \hline

  \begin{tabular}[c]{@{}c@{}}Self-attention in the appearance level,\\ Cross-modal + Self-attention for others\end{tabular} &
  \multicolumn{1}{c|}{2D} &
  \multicolumn{1}{c|}{3D} &
  RoI &
  5.7 &
  16.6 &
  24.2 &
  \textbf{54} \\ \hline

  \begin{tabular}[c]{@{}c@{}}Self-attention in the appearance level,\\ Cross-modal + Self-attention for others\end{tabular} &
  \multicolumn{1}{c|}{2D + 3D} &
  \multicolumn{1}{c|}{2D + 3D} &
  2D + 3D &
  \textbf{6} &
  \textbf{16.7} &
  \textbf{24.6} &
  55 \\ \hline
\end{tabular}%
\end{table*}

\subsubsection{Extended Ablations}

The following ablation presents comprehensive insights on visual feature setting and model design choice. The above part shows that our visual feature setting choice, which is the concatenation of 2D and 3D features, brings higher results even on a different model design than our proposed model. Similarly, the below part indicates that our model design choice performs better even using different feature settings than our proposed approach. Average weighting is applied to the models. These results support our two arguments that using self-attention only at the global level with cross-modal and self-attention at the local levels along with concatenated 2D and 3D features brings the best results. 

\begin{table*}[!htb]
\centering
\caption{The above part shows the Ablation on Visual Feature Settings with Another Model Design on YouCook2 Validation Set. The Result Shows that Concatenated 2D and 3D Features Performs the Best.
The below part indicates the Ablation on Model Design Choice with Another Feature Setting on YouCook2 Dataset. The Result Shows that Our Design Choice Performs Better than Having Only Self-attentions.}
\label{tab:yc2_abl_extra}
\begin{tabular}{|cccccccc|}
\hline
\multicolumn{1}{|c|}{}                                                                                                                          & \multicolumn{3}{c|}{\textbf{Visual  Features}}                                                                         & \multicolumn{1}{c|}{}                                & \multicolumn{1}{c|}{}                                & \multicolumn{1}{c|}{}                                 &                                 \\ \cline{2-4}
\multicolumn{1}{|c|}{\multirow{-2}{*}{\textbf{Attention}}}                                                                                      & \multicolumn{1}{c|}{\textbf{Appearance}} & \multicolumn{1}{c|}{\textbf{Action}} & \multicolumn{1}{c|}{\textbf{Object}} & \multicolumn{1}{c|}{\multirow{-2}{*}{\textbf{R@1↑}}} & \multicolumn{1}{c|}{\multirow{-2}{*}{\textbf{R@5↑}}} & \multicolumn{1}{c|}{\multirow{-2}{*}{\textbf{R@10↑}}} & \multirow{-2}{*}{\textbf{MdR↓}} \\ \hline \hline
\multicolumn{8}{|c|}{\cellcolor[HTML]{EFEFEF}\textbf{Visual Feature Setting Ablation on Another Model Design}}                                                                                                                                                                                                                                                                                                                                                                   \\ \hline
\multicolumn{1}{|c|}{Self-attention for all levels}                                                                                             & \multicolumn{1}{c|}{2D}                  & \multicolumn{1}{c|}{2D}              & \multicolumn{1}{c|}{2D}              & \multicolumn{1}{c|}{5.5}                             & \multicolumn{1}{c|}{\textbf{16.3}}                   & \multicolumn{1}{c|}{24.1}                             & 58                              \\ \hline
\multicolumn{1}{|c|}{Self-attention for all levels}                                                                                             & \multicolumn{1}{c|}{2D}                  & \multicolumn{1}{c|}{3D}              & \multicolumn{1}{c|}{RoI}             & \multicolumn{1}{c|}{5.3}                             & \multicolumn{1}{c|}{16.2}                            & \multicolumn{1}{c|}{23.1}                             & 57                              \\ \hline
\multicolumn{1}{|c|}{Self-attention for all levels}                                                                                             & \multicolumn{1}{c|}{2D + 3D}             & \multicolumn{1}{c|}{2D + 3D}         & \multicolumn{1}{c|}{2D + 3D}         & \multicolumn{1}{c|}{\textbf{6}}                      & \multicolumn{1}{c|}{16.2}                            & \multicolumn{1}{c|}{\textbf{24.5}}                    & \textbf{55}                     \\ \hline \hline
\multicolumn{8}{|c|}{\cellcolor[HTML]{EFEFEF}\textbf{Model Design Ablation on Another Feature Setting}}                                                                                                                                                                                                                                                                                                                                                                          \\ \hline
\multicolumn{1}{|c|}{Self-attention for all levels}                                                                                             & \multicolumn{1}{c|}{2D}                  & \multicolumn{1}{c|}{3D}              & \multicolumn{1}{c|}{RoI}             & \multicolumn{1}{c|}{5.3}                             & \multicolumn{1}{c|}{16.3}                            & \multicolumn{1}{c|}{23.1}                             & 57                              \\ \hline
\multicolumn{1}{|c|}{\begin{tabular}[c]{@{}c@{}}Self-attention in the appearance level,\\ Cross-modal + Self-attention for others\end{tabular}} & \multicolumn{1}{c|}{2D}                  & \multicolumn{1}{c|}{3D}              & \multicolumn{1}{c|}{RoI}             & \multicolumn{1}{c|}{\textbf{5.7}}                    & \multicolumn{1}{c|}{\textbf{16.6}}                   & \multicolumn{1}{c|}{\textbf{24.2}}                    & \textbf{54}                     \\ \hline
\end{tabular}
\end{table*}

\section{Conclusion}
 
In this work, we propose a novel hierarchical MoE transformer to transform the text and video features into three semantic roles and then align them in three joint embedding spaces. To better use the intra-modality and inter-modality video features, we employ self-attention and cross-modal attention at the local levels and only use self-attention at the global level. Moreover, the weighted sum of the video features boosts the results by exploiting the discriminative video features. Our approach surpasses SOTA methods on the YouCook2 and MSR-VTT datasets by using the same visual feature set without pre-training. We also share an extensive ablation study allowing us to have deeper insights. 

\section*{Acknowledgments}
This research is supported by the Agency for Science, Technology and Research (A*STAR) under its AME Programmatic Funding Scheme (Project A18A2b0046).

\bibliographystyle{IEEEtran}
\bibliography{IEEEabrv,refs}

\newpage

\section{Biography Section}

\vspace{1pt}

\begin{IEEEbiography}[{\includegraphics[width=1in,height=1.25in,clip,keepaspectratio]{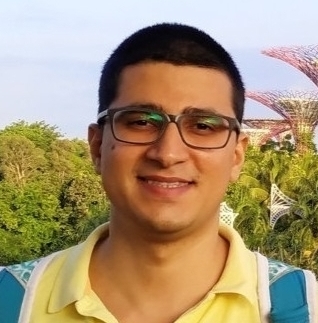}}]{Burak Satar} is a third-year Ph.D. candidate majoring in video understanding with multi-modal features in a joint program at the School of Computer Science and Engineering Nanyang Technological University, Singapore and Institute for Infocomm Research, A*STAR Singapore, under SINGA scholarship. He received his M.Sc. degree in Electronic Engineering in 2018. His research interests include causality in vision, multimedia retrieval, and computer vision.
\end{IEEEbiography}

\begin{IEEEbiography}[{\includegraphics[width=1in,height=1.25in,clip,keepaspectratio]{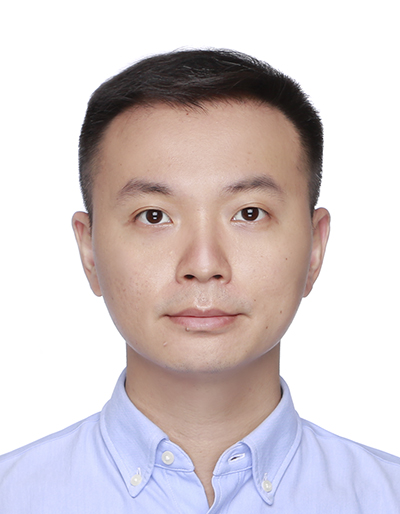}}]{Hongyuan Zhu} received the B.S.degree in software engineering from the University of Macau, in 2010, and the Ph.D. degree in computer engineering from Nanyang Technological University, Singapore, in 2014. He is currently a Research Scientist with the Institute for Infocomm Research, A*STAR, Singapore. His research interests include multimedia content analysis and segmentation.
\end{IEEEbiography}

\begin{IEEEbiography}[{\includegraphics[width=1in,height=1.25in,clip,keepaspectratio]{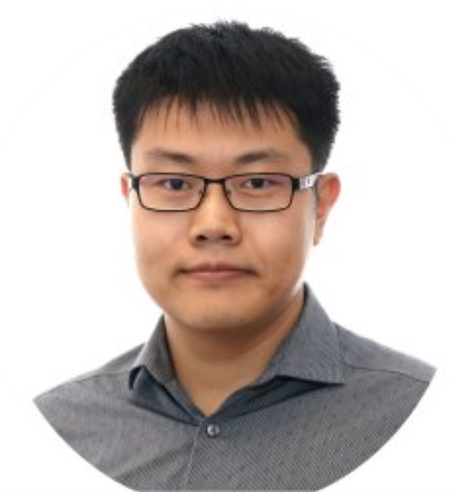}}]{Hanwang Zhang} is currently an Assistant Professor at the School of Computer Science and Engineering, Nanyang Technological University, Singapore. He was a research scientist at the Department of Computer Science, Columbia University, USA. He received the B.Eng (Hons.) degree in computer science from Zhejiang University, Hangzhou, China, in 2009, and the Ph.D. degree in computer science from the National University of Singapore in 2014. His research interest includes computer vision, multimedia, and social media. Dr. Zhang is the recipient of the Best Demo runner-up award in ACM MM 2012, the Best Student Paper award in ACM MM 2013, the Best Paper Honorable Mention in ACM SIGIR 2016, and TOMM best paper award 2018. He is also the winner of Best Ph.D. Thesis Award of School of Computing, National University of Singapore, 2014.
\end{IEEEbiography}

\begin{IEEEbiography}[{\includegraphics[width=1in,height=1.25in,clip,keepaspectratio]{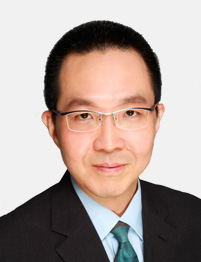}}]{Joo Hwee Lim} received his B.Sc. (1st Class Honours) and M.Sc. (by research) degrees in Computer Science from the National University of Singapore and his Ph.D. degree in Computer Science \& Engineering from the University of New South Wales. He joined Institute for Infocomm Research (I2R), Singapore (and its predecessors) in October 1990. His research experience includes connectionist expert systems, neural-fuzzy systems, handwritten recognition, multi-agent systems, content-based image retrieval, scene/object recognition, and medical image analysis, with over 280 international refereed journal and conference papers and 25 patents (awarded and pending).
He is currently Principal Scientist and Department Head (Visual Intelligence) at I2R, A*STAR, Singapore, and an Adjunct Professor at the School of Computer Science and Engineering, Nanyang Technological University, Singapore. He was also the co-Director of IPAL (Image, Pervasive Access Lab), a French-Singapore Joint Lab (UMI 2955, Jan 2007 to Jan 2015). He was bestowed the title of 'Chevalier dans l'ordre des Palmes Academiques' by the French Government in 2008 and the National Day Commendation Medal by the Singapore Government in 2010.
\end{IEEEbiography}

\end{document}